\documentclass[runningheads]{llncs}

\usepackage[year=2026,ID=5222]{eccv}

\usepackage{eccvabbrv}
\usepackage{graphicx}
\usepackage{booktabs}
\usepackage{amsmath}
\usepackage{amssymb}
\usepackage{multirow}
\usepackage[accsupp]{axessibility}
\usepackage{xcolor}

\usepackage[pagebackref,breaklinks,colorlinks,citecolor=eccvblue]{hyperref}
\usepackage{orcidlink}

\newcommand{\naming}{\textbf{ByteLOOM}}
\newcommand{\namingplain}{ByteLOOM}
\newcommand{\namingbm}{\textbf{Mani4D}}

\newcommand{\redcite}[1]{{\cite{#1}}}

\begin{document}

\title{\naming: Weaving Geometry-Consistent Human-Object Interactions Videos through Progressive Curriculum Learning}

\titlerunning{\naming}

\author{Bangya Liu\inst{1,\dag} \and
Xinyu Gong\inst{2} \and
Zelin Zhao\inst{3,\dag} \and
Ziyang Song\inst{4,\dag} \and
Yulei Lu\inst{2} \and
Suhui Wu\inst{2} \and
Jun Zhang\inst{2} \and
Suman Banerjee\inst{1} \and
Hao Zhang\inst{2}}

\authorrunning{B.~Liu et al.}

\institute{University of Wisconsin-Madison \and
ByteDance \and
Georgia Institute of Technology \and
The Hong Kong Polytechnic University\\
\textsuperscript{\dag}Project during internship at ByteDance}

\maketitle

\begin{abstract}
Human-object interaction (HOI) video generation has garnered increasing attention due to its promising applications in digital humans, e-commerce, advertising, and robotics imitation learning. However, existing methods face two critical limitations: (1) a lack of effective mechanisms to inject multi-view information of the object into the model, leading to poor cross-view consistency, and (2) heavy reliance on fine-grained annotations including hand mesh and body templates for modeling interaction occlusions, due to limited training data. To address these challenges, we introduce \namingplain, a Diffusion Transformer (DiT)-based framework that generates realistic HOI videos with geometrically consistent object illustration, using simplified human conditioning and 3D object inputs. We first propose an RCM-cache mechanism that leverages Relative Coordinate Maps (RCM) as a universal representation to maintain object's geometry consistency and precisely control 6-DoF object transformations in the meantime. To compensate HOI dataset scarcity and leverage existing datasets, we further design a training curriculum that enhances model capabilities in a progressive style and relaxes the demand of hand mesh. Extensive experiments demonstrate that our method faithfully preserves human identity and the object's multi-view geometry, while maintaining smooth motion and object manipulation. Project page: \href{https://neutrinoliu.github.io/byteloom/}{neutrinoliu.github.io/byteloom/}
\end{abstract}
\vspace{-0.5cm}


\begin{figure}[t]
  \centering
  \includegraphics[width=\textwidth]{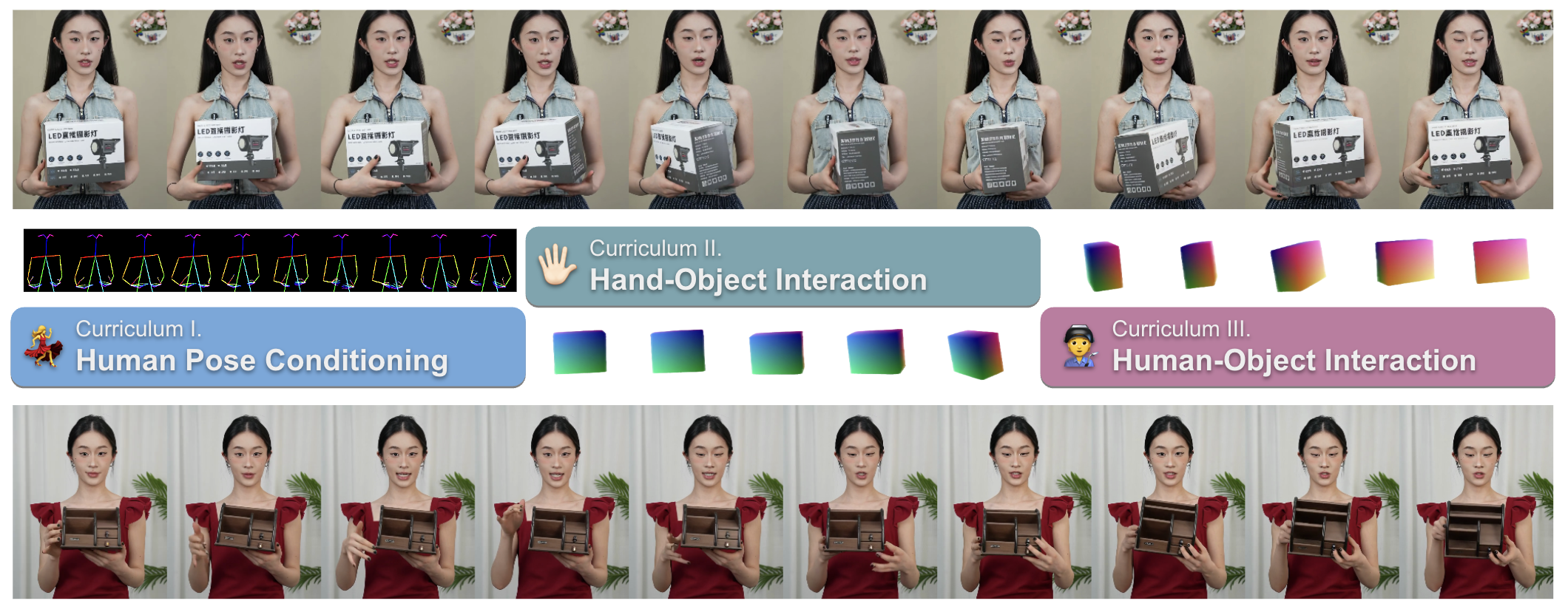}
  \caption{\namingplain{} generates high-quality human-object interaction videos while maintaining consistent object geometry and appearance across frames. Our method introduces two key innovations: First, drawing an analogy to skeleton-based human pose conditioning, we propose Relative Coordinate Maps (RCM) as a universal representation for 3D object poses. Second, we develop a multi-stage training curriculum that enables the diffusion model to learn HOI video generation progressively, maximizing the utility of available datasets.}
  \vspace{-0.5cm}
  \label{fig:teaser}
\end{figure}
\section{Introduction}
\label{sec:intro}

The synthesis of realistic human-object interaction (HOI) videos has emerged as a critical research area with applications spanning digital commerce, entertainment, robotics, and virtual production. While recent advances in video generation have demonstrated remarkable capabilities in synthesizing human motion and appearance~\cite{wang2025unianimate,zhang2024mimicmotion,rd-dit}, generating videos featuring complex object manipulation with geometric consistency remains a significant challenge.

Recent efforts have made progress on specific aspects of this challenge. AnchorCrafter~\cite{anchorcrafter} as the pioneer work introduces a dedicated bundle of conditions, including human reference, human pose, hand mesh, object reference and object depth map, to achieve the HOI rendering. HunyuanVideo-HOMA~\cite{hunyuanhoma} further relaxes the dependency on curated motion data through weakly conditioned multimodal guidance. DreamActor-H1~\cite{dreamactor} employs masked cross-attention and a large 3D motion templates for human-product demonstration videos. Despite these advances, two fundamental limitations persist: (1) the lack of effective mechanisms to inject multi-view information into diffusion models, leading to poor geometric consistency of manipulated objects, and (2) a heavy reliance on fine-grained annotations like hand mesh that are costly to scale up.

To address these challenges, we introduce \naming, a DiT-based framework that generates high-fidelity HOI videos with geometrically consistent object manipulation using simplified human conditioning and 3D object inputs.

The first innovation of our approach is a three-stage curriculum learning strategy that gradually builds the model's capability from basic human motion understanding to full human-object interaction. Because the availability of training data varies dramatically across interaction complexity, we structure the curriculum to exploit this imbalance.

By progressing from widely available pose-only data, to plentiful hand-object data, and finally to scarce full-interaction data, the curriculum equips the model with strong foundational motion understanding before introducing increasingly complex manipulation tasks. This staged design leads to better generalization, more stable learning, and significantly more consistent HOI synthesis.

Our second innovation is the RCM (Relative Coordinate Map) cache, designed to provide the diffusion model with a persistent and reliable 3D prior. Instead of relying on single-frame or one-time conditioning, we bind multi-view reference RGB images with their corresponding RCMs and store them as a cache accessible to the DiT throughout generation. During synthesis, the model also receives per-frame RCM inputs that specify the object's target 6-DoF pose and geometry for that timestep. By jointly leveraging the cached multi-view RCM-RGB pairs as a global appearance prior and the per-frame RCM as a geometric control signal, the model implicitly learns to retrieve the correct object appearance from the cache and render it faithfully in every frame. This mechanism ensures that the object remains geometrically consistent, correctly textured, and view-aligned even under complex manipulation.

Extensive experiments demonstrate that our method preserves human identity, maintains smooth motion, and substantially surpasses prior work in object manipulation fidelity and multi-view consistency. Overall, our contribution could be summarized as follows:
\begin{itemize}
\item We design a structured training curriculum---comprising human-pose pretraining, hand-object pretraining, and HOI finetuning---that effectively balances the scarcity of high-quality HOI data with the progressively increasing granularity of interaction conditioning.
\item We introduce RCM-cache, a universal mechanism for injecting multi-view 3D object information and precisely controlling object 6-DoF transformations. This persistent 3D prior enables substantially improved geometric consistency across video frames.
\item We designed and practiced a depth-free data curation pipeline that robustly extracts HOI conditionings including human pose and object 6DoF pose.
\item We establish \namingbm, a dedicated benchmark for evaluating geometry-consistent object manipulation in generated HOI videos, enabling systematic and quantitative comparison across methods. Benchmark and finetune dataset will be released in the camera ready version.
\end{itemize}

\section{Related Work}

\subsection{Human-Object Interaction Video Generation}

Human-object interaction (HOI) generation has emerged as a critical challenge in creating physically plausible visual content.

In the \textbf{motion synthesis} domain, extensive research has focused on predicting human motions during object interactions, with methods like InterDiff~\cite{xu2023interdiff}, TeSMo~\cite{yi2024generating}, ROG~\redcite{rog}, and HOIAnimator~\cite{song2024hoianimator} generating interaction sequences based on 3D object representations. Those methods show great potential of providing precise human pose and object pose conditioning for pixel level interaction generation.

For \textbf{image generation}, InteractDiffusion~\cite{hoe2024interactdiffusion} provides interaction controllability through text-to-image diffusion models, while VirtualModel~\cite{chen2024virtualmodel} synthesizes HOI images by leveraging input objects and human poses. HOGAN~\cite{hu2022hand} addresses hand-object interactions via split-and-combine approaches considering inter-occlusion patterns. Other generalized methods such as~\cite{lapflow} could also enhance the quality of the generation.

Within \textbf{video generation}, two primary paradigms have emerged: \textbf{(a) video editing} and \textbf{(b) generation from scratch}. Editing-focused methods like HOI-Swap~\cite{xue2024hoi} and ReHoLD~\cite{fan2025ReHOLD} enable object replacement in existing hand-centric videos while preserving interaction realism, whereas MIMO~\cite{men2025mimo} and AnimateAnyone~2~\cite{hu2025animate} substitute human subjects in complex dynamic scenes. However, these approaches remain constrained to editing existing content and often introduce visual artifacts when handling complex interactions occupying small pixel regions. For generation from scratch, AnchorCrafter~\cite{anchorcrafter} integrates HOI into pose-guided frameworks using multi-conditional inputs such as pose and depth, while ManiVideo~\cite{pang2025manivideo} achieves precise control through 3D modeling of both human hands and objects. DreamActor-H1~\cite{dreamactor} extends this route with a large library consisting numerous mesh-based templates. Nevertheless, these methods typically rely on strong conditional inputs or lack sufficient generative freedom, highlighting the need for approaches that balance controllability with generalizability across diverse objects and interaction scenarios while maintaining high-quality synthesis of full-body human-object interactions. HunyuanVideo-HOMA~\cite{hunyuanhoma}, introducing a totally different paradigm, adopts weak condition to relax this tension yet fails to display diverse viewpoints of the object.

\subsection{Geometry-Consistent Generation}

Geometry-consistent generation aims to preserve the underlying 3D structure, shape, and texture of objects across viewpoints and over time.

A first line of work focuses on multi-view image generation as a strong 3D prior~\cite{shi2023mvdream,he2024lucidfusion,GTK}. MVDream~\cite{shi2023mvdream} trains a multi-view diffusion model on both 2D images and 3D data to produce sets of images that are coherent across viewpoints, combining the diversity of 2D diffusion with the cross-view consistency of 3D renderings. LucidFusion~\cite{he2024lucidfusion} similarly targets multi-view coherence by converting each input image into a Relative Coordinate Gaussian representation and using differentiable 3D Gaussian rendering to enforce globally aligned geometry without requiring camera poses. 
More recently, several works extend such \textbf{3D-aware priors} to videos or directly model geometry-consistent video generation. Zhang \emph{et al.} propose 4Diffusion~\cite{zhang20244diffusion}, which augments a 3D-aware image diffusion backbone with temporal attention layers to synthesize view-consistent videos. Other approaches incorporate explicit 3D control signals into diffusion models to tie video frames to stable underlying geometry. Diffusion-as-Shader~\cite{gu2025diffusion} conditions frame synthesis on tracked 3D point trajectories and uses color anchors to bind RGB appearance to persistent 3D points, leading to stable surface textures over time. Related methods leverage object depth maps or pseudo-3D cues~\cite{mvs2v, wang2025cinemaster,anchorcrafter,cetcam} to guide object placement, improving multi-view and temporal consistency while still relying on 2D video backbones.

\section{Method}
\label{sec:method}
\begin{figure}[t]
  \centering
  \includegraphics[width=1.0\textwidth]{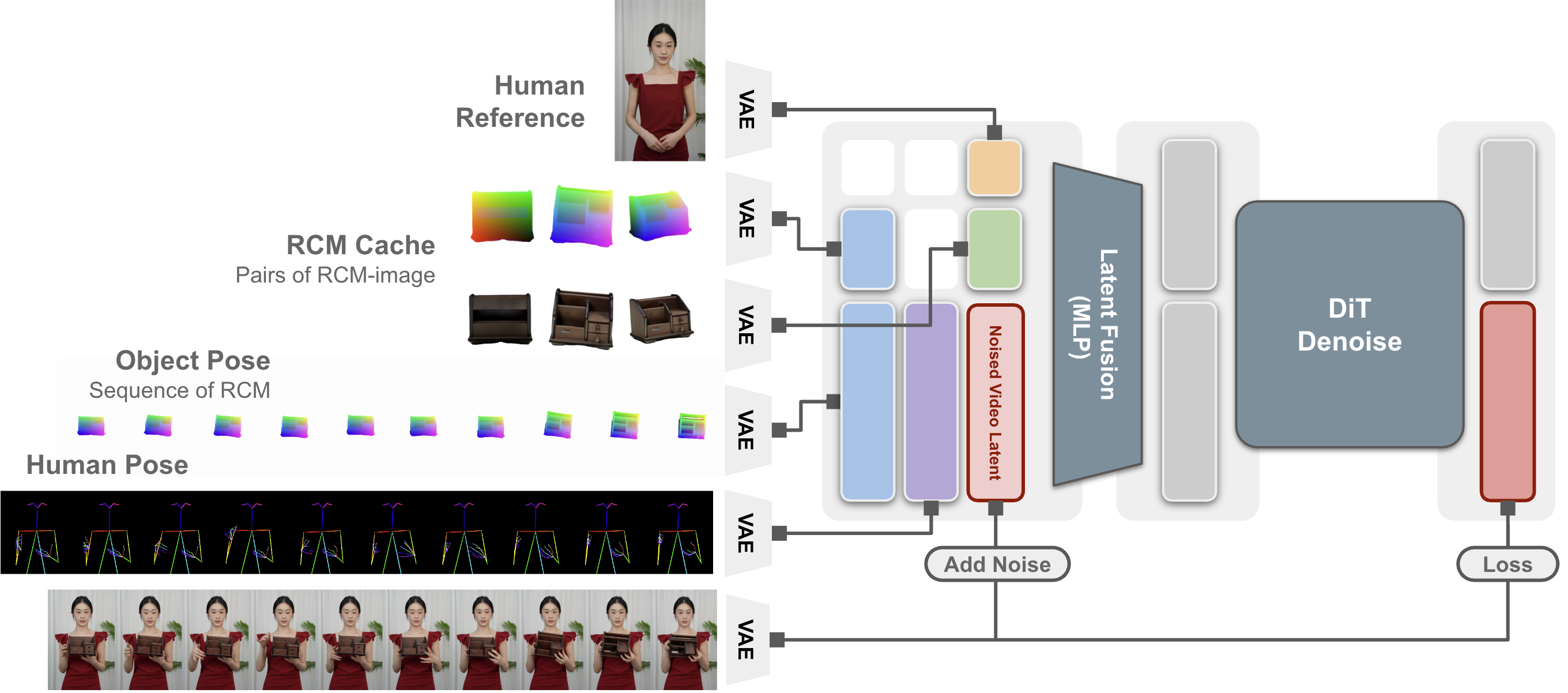}
  \caption{Overview of \naming~training strategy. Different kinds of conditions are concatenated into input latent. Then pass through a latent fuser implemented as a MLP.}
  \vspace{-.5cm}
  \label{fig:overview}
\end{figure}
\subsection{Overview}
Following pioneered works~\cite{hunyuanhoma,dreamactor,anchorcrafter,cetcam}, \naming{} takes inputs condition of both human and objects and generates realistic human-object interaction video as outputs. For the conditioning injection, inspired by the insights from~\cite{rd-dit,cetcam}, we plainly concat and inject all conditions through the input latent fusion, where we let the self-attention to learn the relationship between those input conditions and the final output, as long as we have provided all required information. \cref{fig:overview} shows the overview of our system.

Compared with previous works, \naming{} distinguishes itself from two perspectives: (1) instead of a front-view dominant object presentation, we aim to present a much wider range of views for the object. This raises the requirement of an effective mechanism to deliver such multi-view information into the generation process. (2) instead of relying on precise hand mesh as the interaction conditioning~\cite{anchorcrafter,dreamactor}, we merely use the simple OpenPose~\cite{openpose} style skeleton to condition the human pose. Yet this makes it difficult for the model to directly infer the hand-object occlusion relationship via stick figure.

To tackle with the first challenge, we innovatively introduce RCM-cache module, detailedly introduced in \cref{sec:rcmcache}. To better teach the model how to draw realistic interaction frames, we developed a three stage training curriculum, which will be detailed introduced in \cref{sec:curriculum}. Further, we propose a new metric in \cref{sec:geocon} to quantify the geometry consistency of the presented object in the generated video. Finally, we share our experience when we curate our own HOI datasets in \cref{sec:curation}.

\subsection{Universal Object 6DoF Conditioning}
\label{sec:rcmcache}

Previous video generation tasks, either subject-to-video~\cite{liu2025phantom} or human-object interaction video~\cite{hunyuanhoma,dreamactor}, dominantly apply single image as the object conditioning. AnchorCrafter~\cite{anchorcrafter}, though incorporate three images as object conditions, yet all of them are within a very narrow view cone within approximately 30 degrees, which still dramatically constrains the view angles of the object in the generated video. This conditioning paradigm works for simple interactions like plainly moving the object from left to right. Yet for more complex interactions like rotation, inverting, and manipulating, they fail the task inevitably. One of the key reasons is the lacking of multiview information, in which case model could only guess how to draw the unknown face.

\begin{figure}[t]
  \centering
  \begin{subfigure}[b]{0.50\textwidth}
    \centering
    \includegraphics[width=\textwidth]{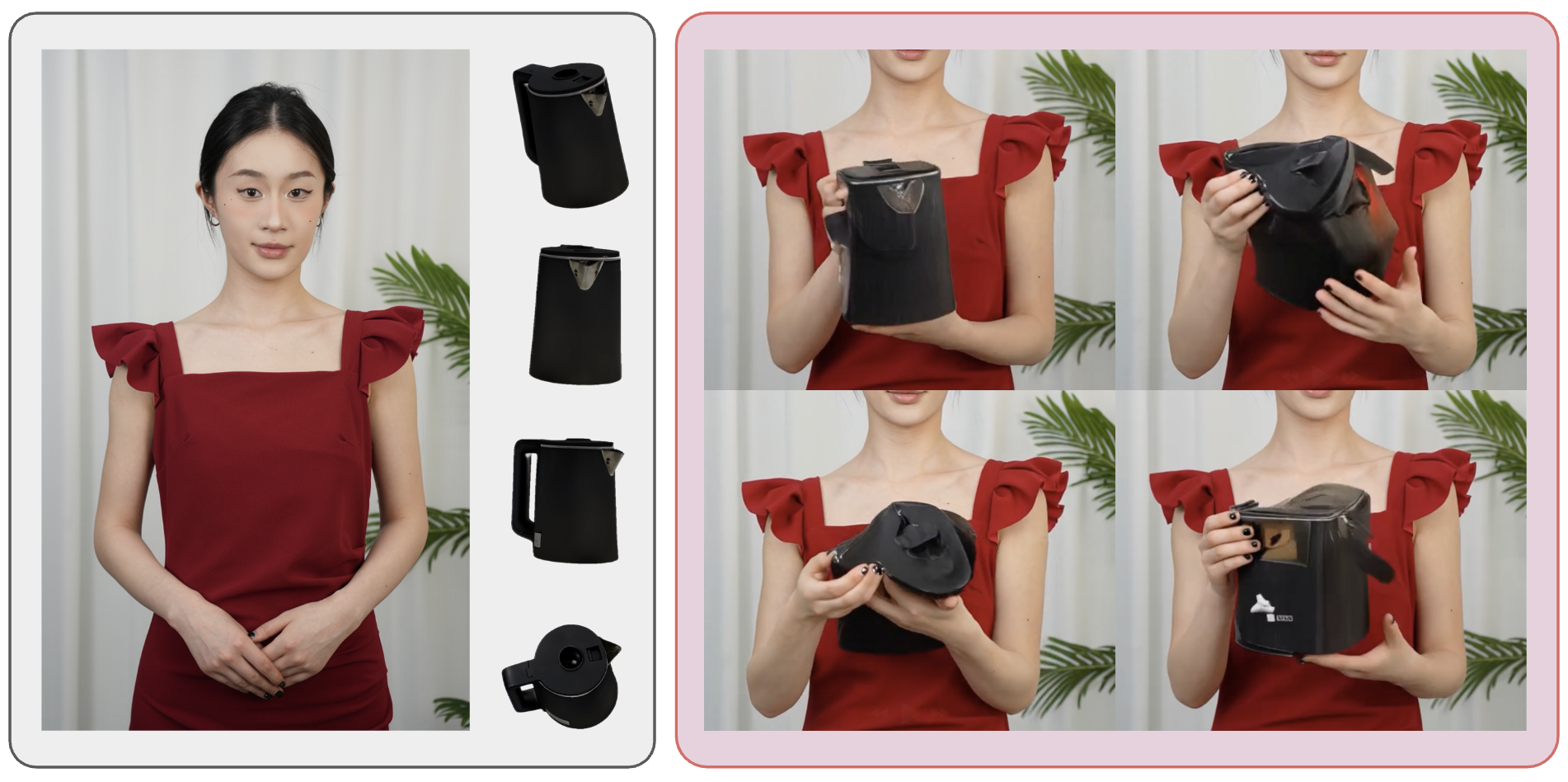}
    \caption{}
    \label{fig:failure}
  \end{subfigure}
  \hfill
  \begin{subfigure}[b]{0.42\textwidth}
    \centering
    \includegraphics[width=\textwidth]{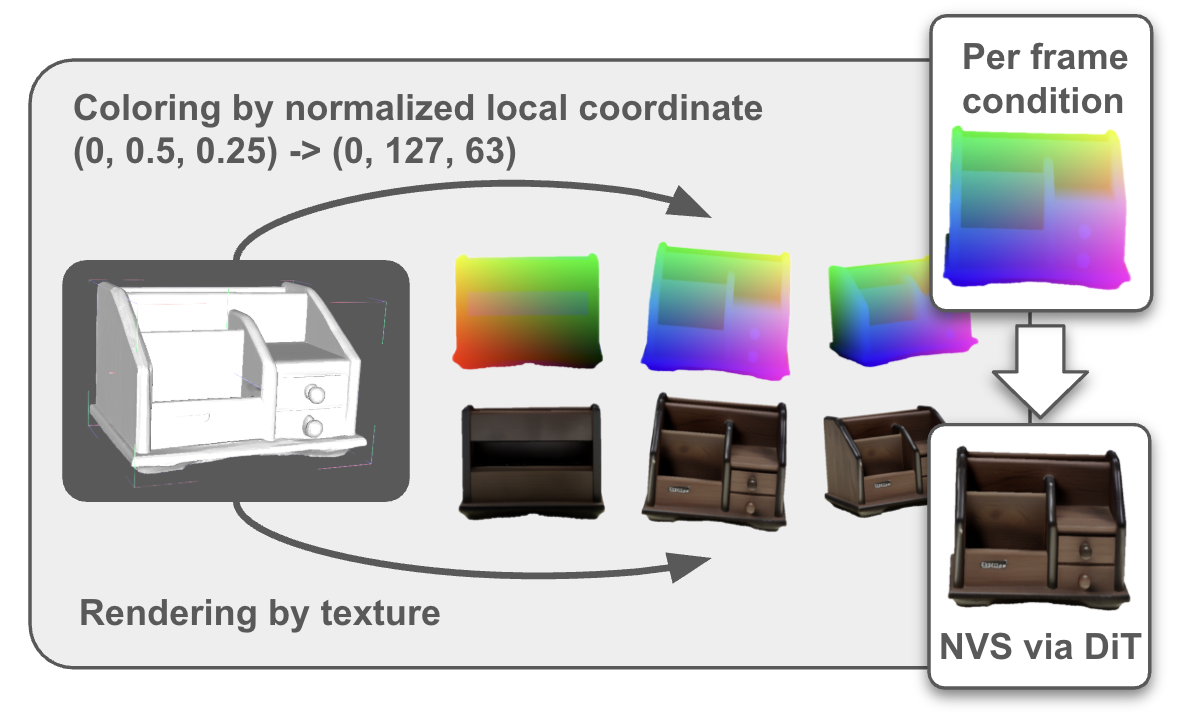}
    \caption{}
    \label{fig:rcmcache}
  \end{subfigure}
  \caption{(a) Model still struggles for generating geometry consistent object, even when multiview object references are provided. (b) RCM-cache of \naming. Sparse views are provided to the model as a reference when performing novel view synthesis.}
  \vspace{-0.5cm}
  \label{fig:failure-and-rcmcache}
\end{figure}

A naive remedy for this problem will be providing multi-view images as a more comprehensive reference. Yet as shown in \cref{fig:failure}, geometry shape of the objects gradually gets twisted and collapse throughout frames, especially when the multi-view images are provided in a random sequence. This could be attributes to that the model does not have the concepts of geometry consistency of a single object, instead it tackles the multiple reference images as multiple independent objects and trying to interpolate them across frames to fulfill the reference conditioning. The remedy here is to construct a strong association in between the multi-view reference images so that the model realize that those reference frames belongs to the same object.

To achieve this, we innovatively introduce relative coordinate map (RCM) into the object conditioning. RCM is originally used in~\cite{he2024lucidfusion} as an auxiliary task for 3D assets generation, which helps to enforce the correspondence between the predicted Gaussian parameters across different views. In \naming, we adopt it as our condition inputs as an explicit 3D prior injection.

As shown in \cref{fig:rcmcache}, to render a relative coordinate map (RCM), geometric surface information is encoded by normalizing vertex positions relative to the object's spatial extents. Specifically, given a vertex position $V_i \in \mathbb{R}^3$ and the object's axis-aligned bounding box defined by two corners $b_{\min}$ and $b_{\max}$ in any body diagonal, the normalized coordinate vector $C_i^{\text{RCM}}$ is calculated via component-wise division:
\begin{equation}
C_i^{\text{RCM}} = \frac{V_i - b_{\min}}{b_{\max} - b_{\min}} \in [0, 1.0]^3
\end{equation}
\begin{equation}
c_i = \text{round}(255 \times C_i^{\text{RCM}}) \in [0, 255]^3
\end{equation}

During the rasterization stage, the rasterizer linearly interpolates the colors defined at the triangle vertices across the surface. For a specific pixel point in the triangle, the final color $c_{\text{pixel}}$ is determined by the barycentric weights $\alpha, \beta, \gamma$ of the triangle:
\begin{equation}
c_{\text{pixel}} = \alpha c_i + \beta c_j + \gamma c_k
\end{equation}

This ensures that every pixel in the output map represents the exact spatial coordinate of the visible surface at that point.

When it comes to video generation, we first construct a RCM-Cache, as shown in \cref{fig:rcmcache}, which is a bunch of RCM and textured images pairs rendered from sparse view angles. We concat this pairs into the frame dimension of video latent as the object reference. Further, for each frame, we render out the desired RCM images with given 6DoF pose, then concat this per frame object pose condition into the channel dimension as shown in \cref{fig:overview}. In this case, we reduce the tough novel view synthesis (NVS) task into a relatively easy texturing task.

One desired property of RCM is such explicitly correspondence in between cross view pixels, so that model is always aware of pixel correlation across frames. Another good property of RCM based object condition is its universality that any object mesh with any texture could be uniformly reduced to a texture look up problem. Even applicable to articulable object as long as there is a consistent mapping between the 3D vertex and its texturing. 

\subsection{Staged Curriculum Learning}
\label{sec:curriculum}

\begin{figure}[t]
  \centering
  \includegraphics[width=1.0\textwidth]{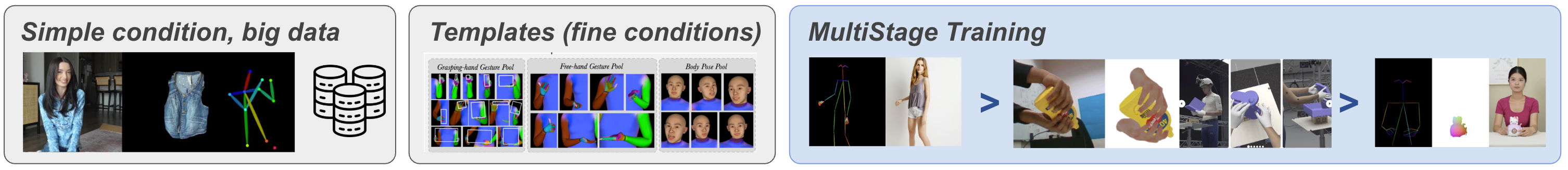}
  \caption{Difference between the training paradigm among HunyuanVideo-HOMA~\cite{hunyuanhoma} (simple condition + data scaling up), DreamActor-H1~\cite{dreamactor} (fine-grained conditioning), and ours (multistage training).}
  \vspace{-0.5cm}
  \label{fig:paradigms}
\end{figure}

We have observed two distinct trends in the previous hand conditioning as shown in \cref{fig:paradigms}: exampled by AnchorCrafter~\cite{anchorcrafter} and DreamActor-H1~\cite{dreamactor} who heavily rely on strong template based conditioning, and a totally different method exampled by HunyuanVideo-HOMA~\cite{hunyuanhoma} who trying to achieve a scaling up with weak human pose conditioning. Both tracks actually reflect the same challenge for the HOI video generation, that there are very limited open HOI dataset with comprehensive object and human annotation.

Realizing this hard tension, we developed a third route that integrates training data from other computer vision tasks like hand-object interaction and object 6DoF estimation. Through a three stage curriculum learning procedure, we let the model learn how to animate human, how to draw geometry consistent object, and how to draw smooth and natural hand occlusion step by step.

The teaser \cref{fig:teaser} shows the curriculum I, II, III that we let the model go through in order to pick up different conditioning capabilities:
\begin{itemize}
\item \textbf{Curriculum~I}: focuses on human-pose-conditioned training, using abundant and easily accessible human motion datasets. This stage teaches the model to generate natural human body movement, temporal dynamics, and coarse spatial structure without involving object interactions.
\item \textbf{Curriculum~II}: leverages the relatively rich hand-object interaction datasets, enabling the model to learn fine-grained skills such as hand-object contact reasoning, grasp configurations, and precise manipulation signals.
\item \textbf{Curriculum~III}: built on the much rarer full human-object interaction datasets, introduces whole-body coordination, global motion patterns, and long-range spatiotemporal consistency under object manipulation.
\end{itemize}

\subsection{Quantify Geometry Consistency}
\label{sec:geocon}

We also need a metric to quantify the object geometry consistency of the generated videos. Existing metrics like MEt3R~\cite{met3r} only operate on image pairs and incur $O(N^2)$ computation complexity. Rolling frame-by-frame pairing could be a feasible but not perfect practice in this case.

To this end, we propose T-SSIM, a metric that leverages 3D Gaussian Splatting (3DGS) as reconstruction primitive for robust geometry consistency quantification. Specifically, we first crop out object segments from the generated video using SAM2 with uniform white padding. Given $n$ frames $I_0, \ldots, I_n$, E-Rayzer~\cite{erayzer} performs feed-forward 3DGS reconstruction, jointly predicting a set of 3D Gaussians $\mathcal{G}$ and camera parameters including a shared intrinsics $K$ and per-frame extrinsics $E_i$:
\begin{equation}
\mathcal{G}, K, \{E_0, \ldots, E_n\} = \text{E-Rayzer}(I_0, \ldots, I_n)
\end{equation}

The predicted 3DGS integrates multi-view geometry into a unified, continuous representation. We then render the 3D Gaussians back to each frame's viewpoint via differentiable splatting:
\begin{equation}
I'_i = \text{Splat}(\mathcal{G}, K, E_i)
\end{equation}

By comparing the rendered image $I'_i$ with the original object crop $I_i$, we quantify geometry consistency as the temporal average of SSIM (T-SSIM) across all frames:
\begin{equation}
\text{T-SSIM} = \frac{1}{n}\sum_{i}\text{SSIM}(I_i, I'_i)
\end{equation}

\subsection{Depth-Free HOI Data Curation}
\label{sec:curation}

\begin{figure}[t]
  \centering
  \begin{subfigure}[b]{0.52\textwidth}
    \centering
    \includegraphics[width=\textwidth]{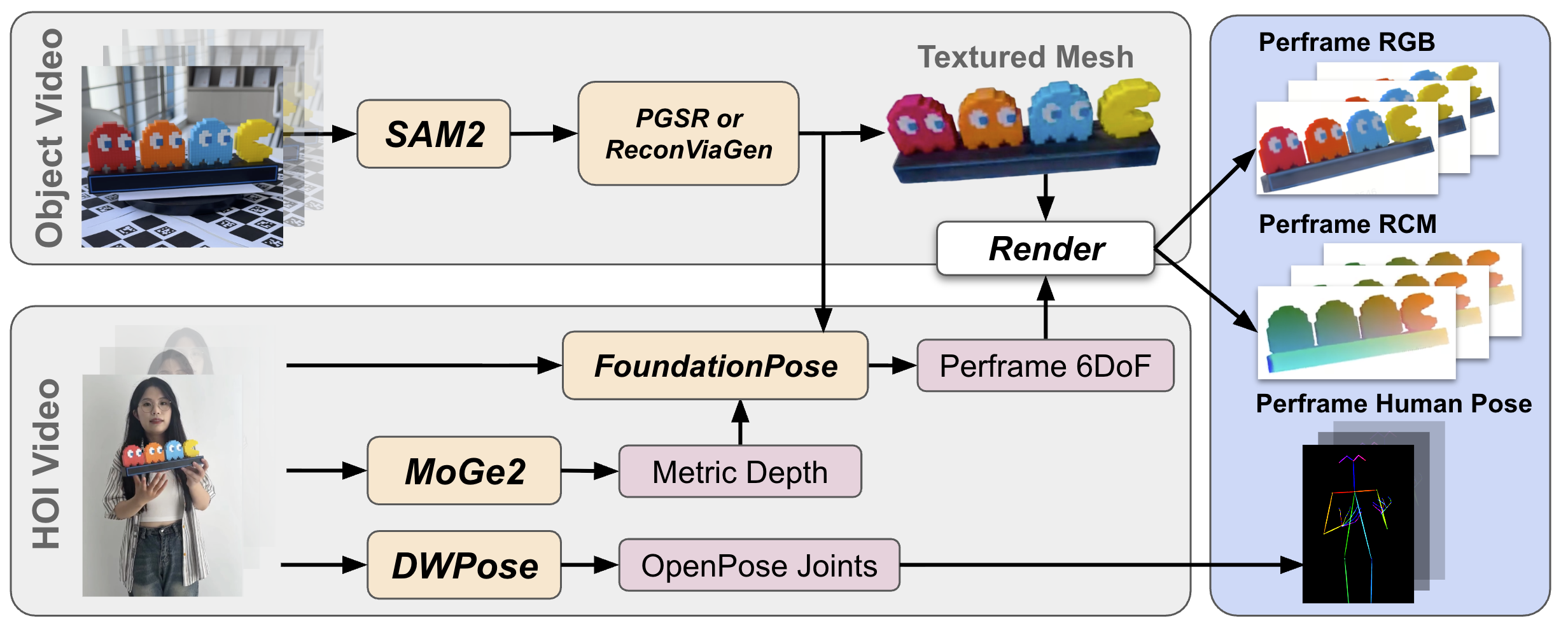}
    \caption{}
    \label{fig:datacuration}
  \end{subfigure}
  \hfill
  \begin{subfigure}[b]{0.44\textwidth}
    \centering
    \includegraphics[width=\textwidth]{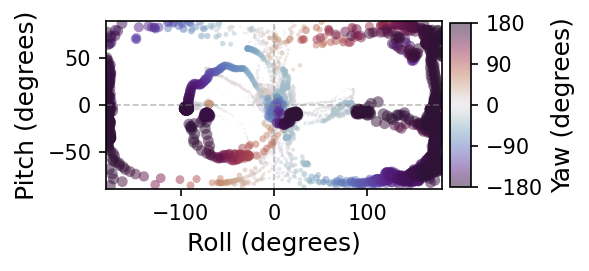}
    \caption{}
    \label{fig:rotation}
  \end{subfigure}
  \caption{(a) Data curation pipeline of \naming. (b) Rotation distribution of \namingbm.}
  \vspace{-0.5cm}
  \label{fig:data_overview}
\end{figure}

The final steps of our curriculum training plan requires a carefully annotated high quality dataset. \cref{fig:datacuration} shows the pipeline of our data curation pipeline for extracting all required conditions. It integrates a wide range of state-of-the-art models to achieve a robust annotation procedure:
\begin{itemize}
\item \textbf{Human Reference}: if a dedicated captured human reference image is not available, we randomly choose one frame from the ground truth video with object masked.
\item \textbf{Human Pose}: we directly leverage DWPose~\cite{dwpose} to infer the per frame human pose, we also filter out joints that is of too low confidence.
\item \textbf{Object Textured Mesh}: if provided with a dedicated object oriented video, first we extract object segments from the original frame sequence via SAM2, similar to the practice in SeconGS~\cite{yin2025semantic}. Then we could directly infer the object mesh via an optimization based method like PGSR~\cite{chen2024pgsr} or 2DGS~\cite{2dgs}. We could also use diffusion based methods like ReconViaGen~\cite{chang2025reconviagen} to acquire a manifold mesh. 
\item \textbf{Metric Depth and Scale}: it could be hard and costly to acquire high quality depth map from the wild video. As an alternative, we adopt MoGe2~\cite{moge2} together with camera intrinsic information extracted from image EXIF tag. Then we plainly align those predicted metric depth with RANSAC to ensure a consistent depth estimation across frames. 
We implant an extra scale calibration module to calibrate the scale of object mesh, inspired by OnePoseViaGen~\cite{oneposeviagen} and RGBTrack~\cite{guo2025rgbtrack}. \item \textbf{Object 6DoF Pose}: Finally we feed the RGB image, depth map as well as the textured object mesh into FoundationPose~\cite{foundationposewen2024} to acquire a robust object 6DoF estimation. one of the key strengths of our curated dataset is its ``multi-view'' nature. \cref{fig:rotation} visualizes the diverse distribution of object rotations across frames relative to the first frame, demonstrating broad coverage of object viewpoints in our training data.
\item \textbf{Object RCM}: it is relatively easy to generate the relative coordinate map of the mesh as long as the object 6DoF and mesh is given. As introduced in \cref{sec:rcmcache}, we could simply recolor each vertex of the mesh with its relative coordinate, the face coloring is linearly interpolated automatically in most modern shaders.
\end{itemize}

\section{Experiments}

\subsection{Training Details}
\label{sec:trainingdetail}

\begin{table}[t]
  \caption{Composition of training data at different curriculum. Each curriculum consists of different combination of data annotation (P -- Human Pose, H -- Hand Pose, O -- Textured Object Mesh) and helps model to learn different conditioning capacity. *We increase the diversity of human actors via different dressing, hair style, and makeup.}
  \label{tab:train_data}
  \centering
  \resizebox{\textwidth}{!}{%
  \begin{tabular}{l|cccc|ccc}
    \toprule
    Curriculum & Dataset & \#Actors & \#Objects & \#Frames & P & H & O \\
    \midrule
    I. Human Pose & proprietary dataset & $\sim$1000 & 0 & 6.4M & \checkmark & \checkmark & \\
    II. Hand-Object Interaction & DexYCB~\cite{chao2021dexycb}, HO3D~\cite{hampali2021ho}, ARCTIC~\cite{fan2023arctic} & 0, 0, 10 & 20, 20, 11 & 520K, 120K, 720K & & \checkmark & \checkmark \\
    Optional: Object Multiview & ScannedGoogle~\cite{gglobj} & 0 & 1030 & 550K & & & \checkmark \\
    III. Human-Object Interaction & Mani4D-Train & 25* & 104 & 45K & \checkmark & \checkmark & \checkmark \\
    \bottomrule
  \end{tabular}%
  }
  \vspace{-0.5cm}
\end{table}

\cref{tab:train_data} shows the composition of the training data for each curriculum mentioned in \cref{sec:curriculum}. We use Wan2.1~\cite{wan2025} as our base model and implemented input latent fusion for condition injection with a MLP. Specifically, we first train model on a large proprietary dataset with human pose annotation extracted by DWPose~\cite{dwpose}. The dataset includes over 100K video clips with diverse human identities, so that our model could pick up a robust pose following capability. 

Further we trained the model on a smaller yet diverse hand-object interaction dataset, components of which includes DexYCB~\cite{chao2021dexycb}, HO3D~\cite{hampali2021ho}, and ARCTIC~\cite{fan2023arctic}, all of which are annotated with both textured 3D mesh, object pose, as well as high quality hand pose labeling. We rendered the object RCM conditions through the method introduced in \cref{sec:rcmcache}. Through this stage, our model enhances its capability of illustrating the contacts and occlusion between hands and objects. 

We also synthesize an object-only dataset where a rigid object freely yet smoothly drifts within the canvas, rendered from ScannedObject~\cite{gglobj} and backgrounded by random sampled interior room image from IKEA indoor dataset~\cite{ikeadataset}. It formats the conditional video generation task as a novel view synthesis task and we hope it could encourage our model to generate cross-view consistent object renderings. However, we found the yield from this training task is marginal hence it is an optional curriculum in our final setup.

In the last stage, Curriculum~III, we carefully curate a small set of videos captured by professional filming studio, named \textbf{Mani4D-Train}, where actors are showcasing a wide range of daily objects and rotate and flip those objects simultaneously. We derive the textured mesh and infer object 6DoF following the practice mentioned in \cref{sec:curation}. Finally, we finish up our curriculum training via this small human-object interaction dataset with comprehensive conditioning. More training detail could be find in our supplementary material.

\begin{figure}[th!]
  \centering
  \captionof{table}{Quantitative results on Mani4D-Test. MEt3R and T-SSIM evaluate multiview geometry consistency via feature similarity~\cite{met3r} and 3DGS reconstruction~\cite{erayzer} respectively. Lower half shows the curriculum ablation study.}
  \label{tab:maintable}
  \resizebox{\textwidth}{!}{%
  \begin{tabular}{l|cccc|ccc|cc}
    \toprule
    Method & Obj-IoU$\uparrow$ & Obj-CLIP$\uparrow$ & Face-Cos$\uparrow$ & LMD$\downarrow$ & Subj-Cons$\uparrow$ & Back-Cons$\uparrow$ & Mot-Smth$\uparrow$ & MEt3R$\downarrow$ & T-SSIM$\uparrow$ \\
    \midrule
    Ground Truth & -- & -- & -- & -- & -- & -- & -- & 0.0419 & 0.9218 \\
    UniAnimate-DiT~\cite{wang2025unianimate} & 0.4644 & 0.7655 & 0.7920 & 0.3260 & \textbf{0.9634} & \textbf{0.9495} & \textbf{0.9940} & 0.0524 & 0.9101 \\
    MimicMotion~\cite{zhang2024mimicmotion} & 0.4647 & 0.7455 & 0.7391 & 0.2017 & 0.9521 & 0.9343 & 0.9923 & 0.0638 & 0.9010 \\
    AnchorCrafter~\cite{anchorcrafter} & 0.6461 & 0.7355 & 0.5771 & 0.2629 & 0.9571 & 0.9413 & 0.9926 & 0.0552 & 0.9079 \\
    \midrule
    Ours (I + II + III) & \textbf{0.8288} & \textbf{0.9100} & 0.8891 & \textbf{0.1427} & 0.9535 & 0.9289 & 0.9930 & 0.0454 & 0.9147 \\
    \midrule
    I + III & 0.7627 & 0.8829 & 0.8648 & 0.2054 & 0.9499 & 0.9371 & 0.9926 & 0.0569 & 0.9048\\
    I + Obj + III & 0.7689 & 0.8901 & \textbf{0.8952} & 0.1930 & 0.9498 & 0.9287 & 0.9929 & 0.0492 & 0.9105\\
    I + II + Obj + III & 0.7770 & 0.8946 & 0.8947 & 0.1861 & 0.9505 & 0.9287 & 0.9926 & \textbf{0.0429} & \textbf{0.9183} \\
    \bottomrule
  \end{tabular}%
  }

  \vspace{0.5em}

  \begin{minipage}[t]{0.49\textwidth}
    \centering
    \captionof{table}{Quantitative ablation study for RCM.}
    \label{tab:rcmablate}
    \resizebox{\textwidth}{!}{%
    \begin{tabular}{l|ccc|cc}
      \toprule
      Ablation & Obj-IoU$\uparrow$ & Obj-CLIP$\uparrow$ & LMD$\downarrow$ & MEt3R$\downarrow$ & T-SSIM$\uparrow$ \\
      \midrule
      without RCM & 0.6619 & 0.8639 & 0.1878 & 0.0540 & 0.9063 \\
      \midrule
      with RCM & 0.8288 & 0.9100 & 0.1427 & 0.0454 & 0.9147 \\

      \bottomrule
    \end{tabular}%
    }
  \end{minipage}%
  \hfill
  \begin{minipage}[t]{0.47\textwidth}
    \centering
    \captionof{table}{Quantitative result with novel human reference.}
    \label{tab:subtable}
    \resizebox{\textwidth}{!}{%
    \begin{tabular}{l|cccc}
      \toprule
      Method & Obj-IoU$\uparrow$ & Obj-CLIP$\uparrow$ & Face-Cos$\uparrow$ & LMD$\downarrow$ \\
      \midrule
      AnchorCrafter~\cite{anchorcrafter} & 0.2435 & 0.6894 & 0.3700 & 0.5797 \\
      \midrule
      Ours & 0.8212 & 0.8957 & 0.7661 & 0.1808 \\
      \bottomrule
    \end{tabular}%
    }
  \end{minipage}

  \vspace{0.5em}

  \includegraphics[width=0.6\textwidth]{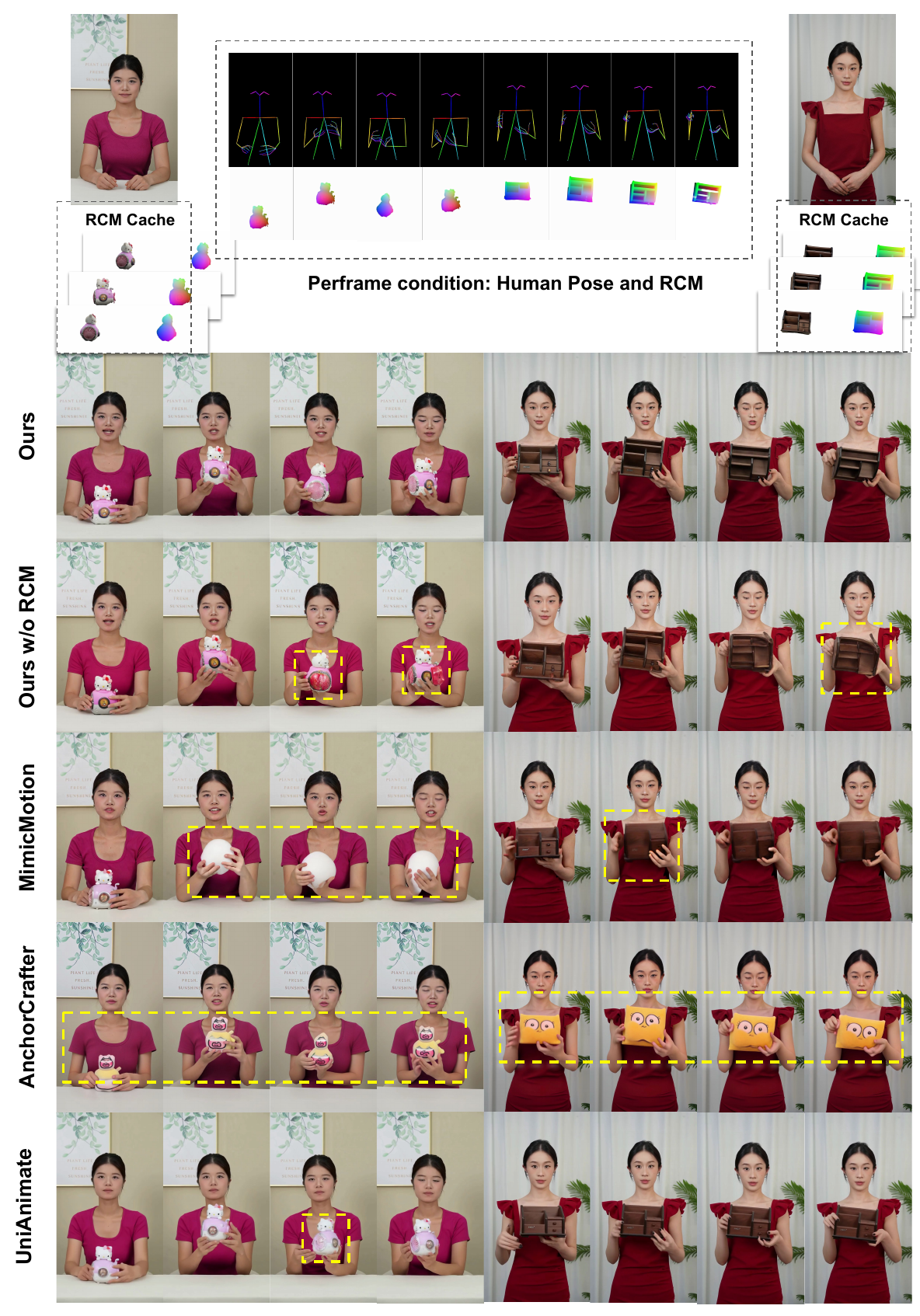}
  \captionof{figure}{Qualitative comparison with prior work.}
  \label{fig:mainvisual}
  \vspace{-0.5cm}
\end{figure}

\subsection{Baseline and Metrics}

Though AnchorCrafter~\cite{anchorcrafter} open-source testset to quantify the model's general capacity of generating Human-Object Interaction videos yet we need a testset that has more rich and abundant object manipulation like rotation and flipping to show the different view point of the object instead of plainly showing the front view, emphasizing the 3D nature of our model. Further the AnchorCrafter dataset does not provide reliable 3D mesh, neither camera intrinsics nor metric depths that are essential for our RCM condition rendering. Under this motivation, we split a small portion from our curated dataset and make it \textbf{Mani4D-Test}. There are totally 5 unique human references and 15 test sequences, with each sequence has a unique object rotating from 90\textdegree{} up to 180\textdegree. 6 objects among them have never shown up in our training set and none of the human-object combination has been shown in training set as well. Each sequence is of 97 frames, approximately 6 seconds under 15 FPS.

We follow the metrics setup as previous works~\cite{anchorcrafter}, integrating metrics including Object-IoU, Object-CLIP for object fidelity, Face-Cos-Similarity, LMD to quantify human and pose observance, and VBench~\cite{vbench} for overall generated video quality.

Aside of all those video generation metrics, we use T-SSIM proposed in \cref{sec:geocon} to quantify the object geometry consistency across the frame.

For baseline, we select AnchorCrafter~\cite{anchorcrafter} as our main baseline, which is the only open-sourced method available in the research community. HunyuanVideo-HOMA~\cite{hunyuanhoma} and DreamActor-H1~\cite{dreamactor} are not publicly accessible, so we also compare with MimicMotion~\cite{zhang2024mimicmotion} and UniAnimate-DiT~\cite{wang2025unianimate} which is a pose driven diffusion model that also based on Wan2.1~\cite{wan2025}. To inject the object information for other baseline methods, we use the first frame of ground truth video where the actor holds the target object. The first frame provides the full object front view and serves as the conditioning of both human and object reference. For AnchorCrafter~\cite{anchorcrafter}, we generate all its required inputs including hand mesh and depth map. Both our method and AnchorCrafter~\cite{anchorcrafter} are provided with human reference image instead of GT first frame.

\subsection{Quantitative Results}

\cref{tab:maintable} reports the quantitative results on \textbf{Mani4D-Test}. Our method surpasses all baselines in object fidelity (Obj-IoU, Obj-CLIP), hand accuracy (LMD), and maintains competitive human similarity (Face-Cos). For overall video quality, we achieve comparable VBench scores with a slight fall back in background consistency, which we attribute to color shifting during multi-segment inference. For multiview geometry consistency, our T-SSIM  closely approaches the ground truth, and we achieve the best MEt3R~\cite{met3r} score among all methods, confirming that our generated objects maintain consistent 3D geometry across frames.

\subsection{Qualitative Results}

We conducted a qualitative comparison with baselines, as shown in \cref{fig:mainvisual}. Our approach produces consistently superior results. MimicMotion~\cite{zhang2024mimicmotion} fails to preserve the geometry of the reference object, as evidenced by distorted shapes highlighted in the bounding boxes. AnchorCrafter~\cite{anchorcrafter} suffers from severe overfitting (SpongeBob characters appearing in generated frames, despite never being conditioned on in our setup) and does not generalize to unseen input references. UniAnimate-DiT~\cite{wang2025unianimate} generates visually plausible videos but cannot faithfully reveal novel viewpoints of the reference object due to the absence of effective multi-view information injection.

\subsection{Ablation Studies}

We show the ablation of our curriculum training in the lower part of \cref{tab:maintable}. Where I, II, Obj, III represents human pose curriculum, hand-object curriculum, optional object NVS curriculum and Mani4D finetuning respectively.

Introducing hand-object dataset brings a guaranteed enhancement for the hand illustration as well as object fidelity. Yet face-cos-similarity seems is of little fluctuation. Adding pure object NVS training task to the curriculum does help the object geometry consistency and full curriculum with object task achieves the best consistency score. However, such yield always comes with a deterioration of the hand interaction illustration. 

Further there is a noticeable performance drop from I+II+III to I+II+Obj+III, which could be attributed for two reasons: first the gaining from pure object curriculum is marginal and could be fully supported by hand-object curriculum alone. Second, the data distribution of our synthetic NVS task is dramatically different from the authentic human object interaction setup, which has the potential to deteriorate the model performance especially the hand fidelity. So in our practice we set pure object NVS task as an optional curriculum.

We also ablate the contribution of RCM mechanism and the result is shown in \cref{tab:rcmablate}. We train a fresh model that uses object multiview images only since Stage~2, providing a fair comparison of the RCM conditioning mechanism. The comprehensive quality enhancement fully demonstrates the effectiveness of RCM mechanism and \cref{fig:mainvisual} also gives a straightforward comparison.

\subsection{Infer on Novel Inputs}

To fully demonstrate that our model learns to generate plausible human-object interaction instead of overfitting \namingbm{} dataset, we also perform video inference in novel inputs, including novel human reference and novel object. Comprehensive comparisons and demo videos could be checked in the project page.

\begin{figure*}[t]
  \centering
  \begin{subfigure}[t]{0.48\textwidth}
    \centering
    \includegraphics[width=\textwidth]{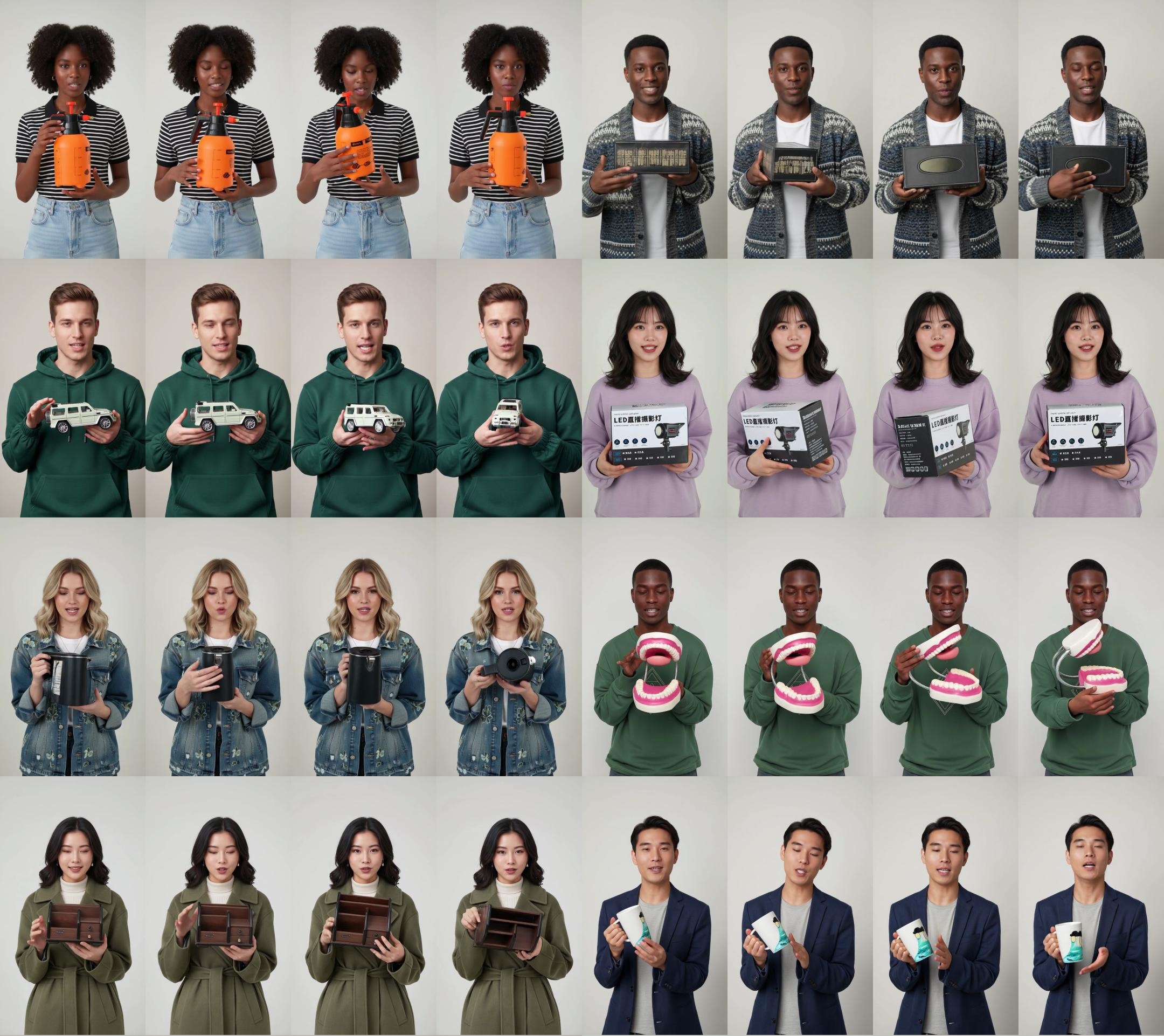}
    \caption{}
    \label{fig:novel_human}
  \end{subfigure}
  \hfill
  \begin{subfigure}[t]{0.48\textwidth}
    \centering
    \includegraphics[width=\textwidth]{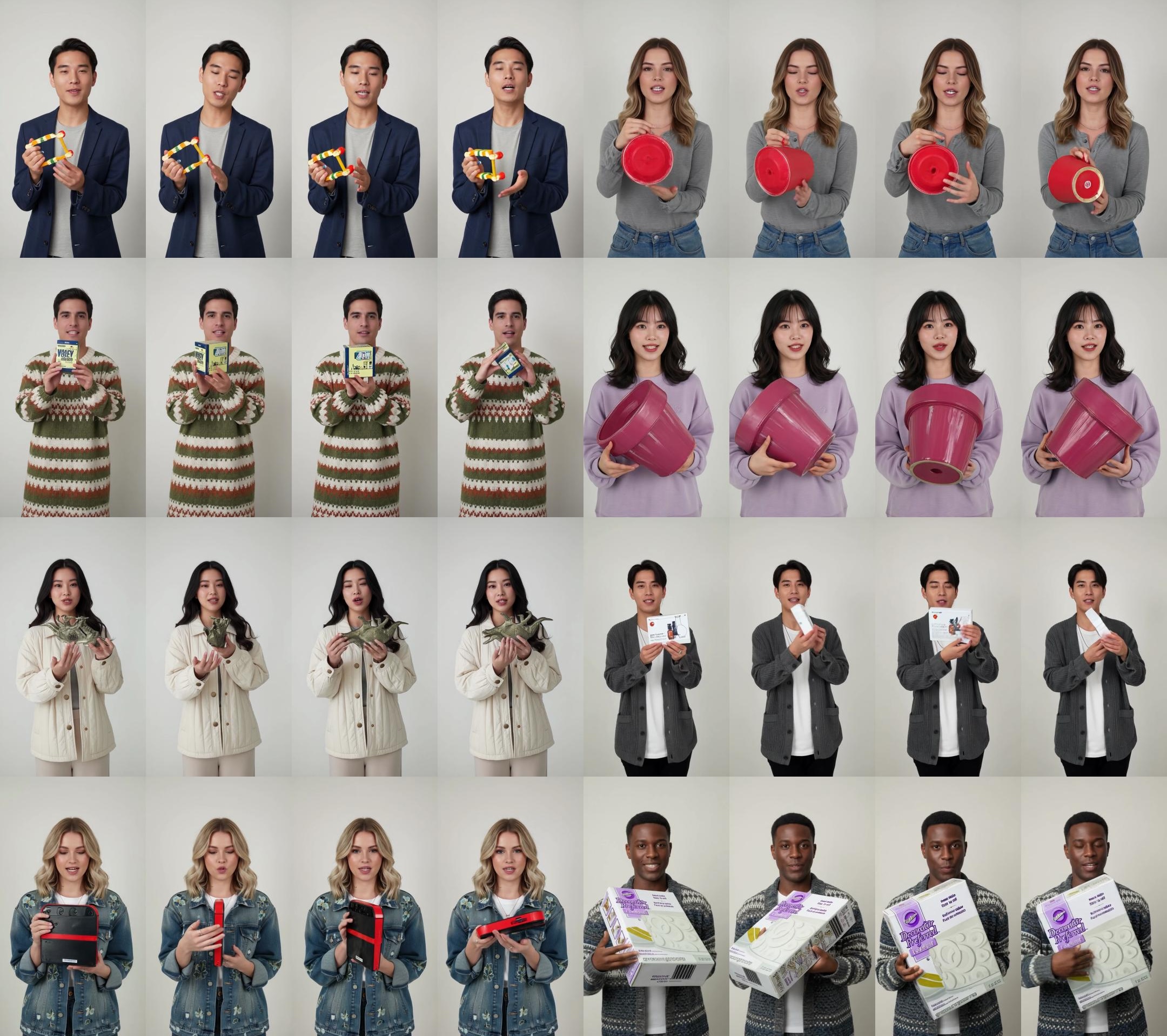}
    \caption{}
    \label{fig:novel_obj}
  \end{subfigure}

  \vspace{0.5em}

  \begin{subfigure}[t]{0.48\textwidth}
    \centering
    \includegraphics[width=\textwidth]{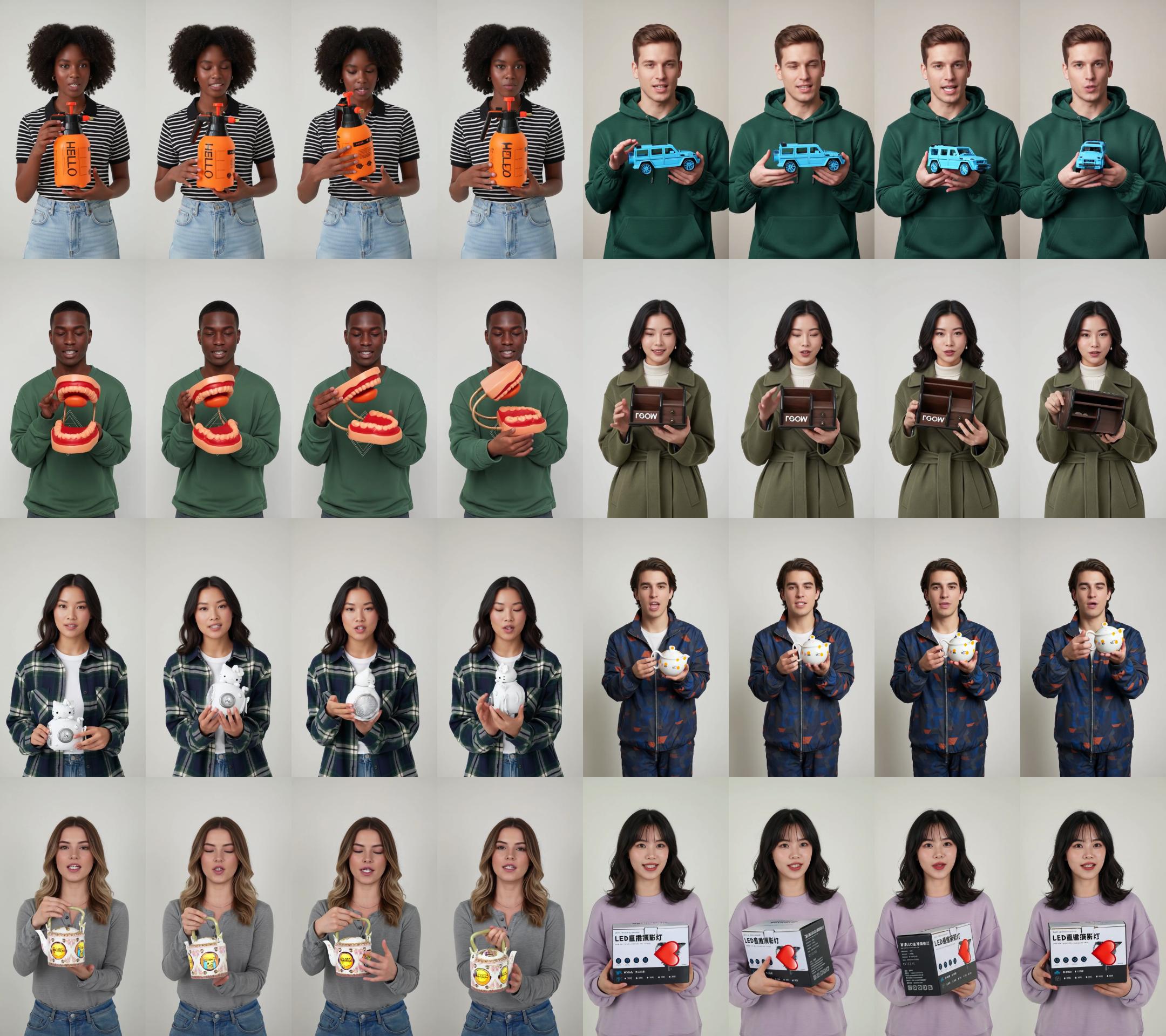}
    \caption{}
    \label{fig:retexture}
  \end{subfigure}
  \hfill
  \begin{subfigure}[t]{0.48\textwidth}
    \centering
    \includegraphics[width=\textwidth]{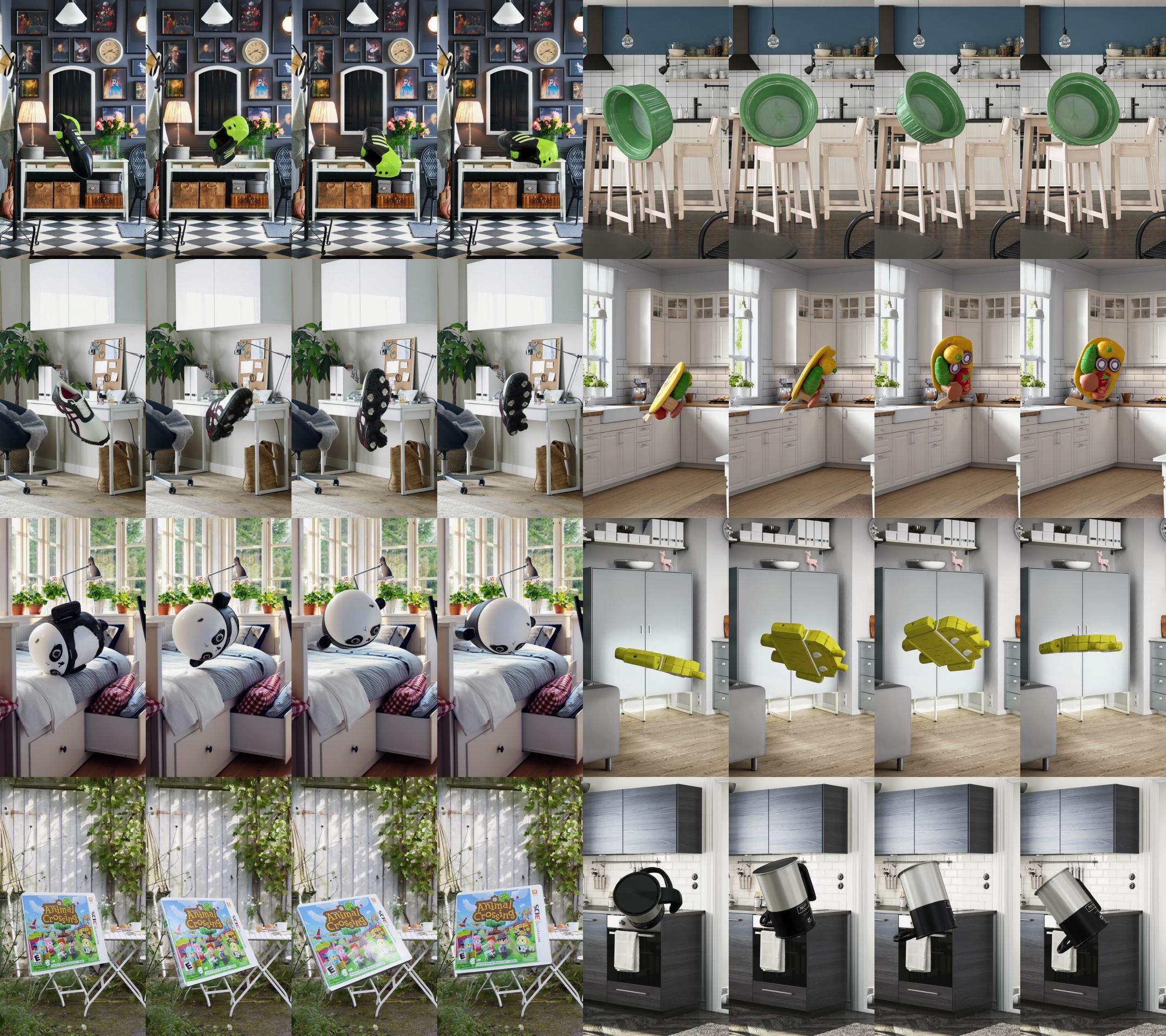}
    \caption{}
    \label{fig:nvs}
  \end{subfigure}
  \caption{Additional qualitative results. (a) Inference with novel human reference. (b) Inference with novel object and novel human reference. (c) \naming~also provide a convenient way for object editing (d) NVS: infer with object-only condition.}
    \vspace{-.5cm}

  \label{fig:additional_results}
\end{figure*}

\textbf{Novel Human.} \cref{tab:subtable} shows the quantitative result when inferring the video on novel human references, all of which are generated by image generation model. \cref{fig:novel_human} provides a qualitative demonstration. From result we could see that our method surpasses AnchorCrafter~\cite{anchorcrafter} even when generating with unseen human reference, meanwhile maintaining a decent hand accuracy (LMD).

\textbf{Novel Object.} We also want to make sure our model could easily generalize to other object and geometry, hence we infer the video with both novel human references and novel objects. 
From \cref{fig:novel_obj}, we could find that our model could successfully illustrate the contact and occlusion even it has never seen the human reference neither the object reference before. Yet when the mismatch between the object geometry and human pose condition is severe, there are obvious penetration effects observed. 

\textbf{Object Texture Manipulation.} Since we condition the object by rendering sparse views of the 3D mesh. We could easily manipulate the object appearance via directly modifying the mesh texture. \cref{fig:retexture} shows such efforts including doodling and color toning. The generated video faithfully reflects the texture modification as we expected.

\textbf{Object Novel View Synthesis.} \cref{fig:nvs} tries to perform the task we previously mentioned as object-only curriculum, yet in a zero-shot style (\ie with no training). The qualitative result successfully demonstrates the novel view synthesis capability facilitated by our RCM mechanism. To our surprise, the object in the generated video could also interact with the background environment in a natural and smooth way (left column in the third row in \cref{fig:nvs}).

\section{Conclusion}

We present \naming, a Diffusion Transformer-based framework for generating geometrically consistent human-object interaction videos. Our approach addresses two critical challenges: limited viewpoint diversity of manipulated objects and dependency on fine-grained hand mesh annotations. Through the RCM-cache mechanism, we achieve precise 6-DoF object control using relative coordinate maps as a universal representation, substantially enhancing cross-frame geometry consistency. Our progressive three-stage training curriculum---encompassing human pose pretraining, hand-object pretraining, and HOI finetuning---effectively overcomes data scarcity while eliminating hand mesh requirements. Combined with our depth-free data curation pipeline, this enables scalable and robust extraction of HOI conditioning signals. We also introduce \namingbm, a new benchmark for evaluating geometry-aware HOI generation capabilities.

\bibliographystyle{splncs04}
\bibliography{main,new_refs}
\end{document}